\documentclass{article}

\usepackage[preprint]{neurips_2025}

\usepackage[utf8]{inputenc} 
\usepackage[T1]{fontenc}    
\usepackage[most]{tcolorbox}

\usepackage[utf8]{inputenc}
\usepackage[T1]{fontenc}

\usepackage{color,xcolor}
\usepackage{epsfig}
\usepackage{graphicx}
\usepackage{duckuments}

\usepackage{adjustbox}
\usepackage{array}
\usepackage{booktabs}
\usepackage{colortbl}
\usepackage{float,wrapfig}
\usepackage{hhline}
\usepackage{multirow}
\usepackage{subcaption}
\usepackage[font=small]{caption}
\usepackage{makecell}
\usepackage{listings}

\usepackage{amsmath,amsfonts,amsthm,amssymb}
\usepackage{bm}
\usepackage{nicefrac}
\usepackage{microtype}
\usepackage[most]{tcolorbox}

\usepackage{changepage}
\usepackage{extramarks}
\usepackage{fancyhdr}
\usepackage{lastpage}
\usepackage{setspace}
\usepackage{soul}
\usepackage{xspace}
\usepackage{indentfirst}
\usepackage{pifont}
\usepackage{cuted}
\usepackage{wrapfig}
\usepackage{threeparttable} 

\usepackage[breaklinks,urlcolor=blue]{hyperref}
\usepackage{url}

\usepackage{algorithm, algorithmic}
\usepackage{enumitem}

\usepackage{wasysym}
\usepackage{todonotes}
\usepackage{pifont}
\usepackage{fancyvrb}
\usepackage{fvextra}
\usepackage{float}
\usepackage{outlines}

\newcolumntype{L}[1]{>{\raggedright\let\newline\\\arraybackslash\hspace{0pt}}m{#1}}
\newcolumntype{C}[1]{>{\centering\let\newline\\\arraybackslash\hspace{0pt}}m{#1}}
\newcolumntype{R}[1]{>{\raggedleft\let\newline\\\arraybackslash\hspace{0pt}}m{#1}}

\newcommand{\sect}[1]{Section~\ref{sect:#1}}
\newcommand{\app}[1]{Appendix~\ref{app:#1}}

\newcommand{\lblsect}[1]{\label{sect:#1}}
\newcommand{\lblapp}[1]{\label{app:#1}}

\newcommand{\ignorethis}[1]{}

\makeatletter
\DeclareRobustCommand\onedot{\futurelet\@let@token\@onedot}
\def\@onedot{\ifx\@let@token.\else.\null\fi\xspace}

\makeatother

\definecolor{citecolor}{rgb}{34,139,34}
\definecolor{mydarkblue}{rgb}{0,0.08,1}
\definecolor{mydarkgreen}{rgb}{0.02,0.6,0.02}
\definecolor{mydarkred}{rgb}{0.8,0.02,0.02}
\definecolor{mydarkorange}{rgb}{0.40,0.2,0.02}
\definecolor{mypurple}{RGB}{111,0,255}
\definecolor{myred}{rgb}{1.0,0.0,0.0}
\definecolor{mygold}{rgb}{0.75,0.6,0.12}
\definecolor{mydarkgray}{rgb}{0.66,0.66,0.66}
\definecolor{cvprblue}{rgb}{0.21,0.49,0.74}
\definecolor{lightblue}{rgb}{0.95,0.95,1}
\definecolor{darkblue}{rgb}{0.2,0.4,0.6}

\definecolor{darkgreen}{rgb}{0.15, 0.75, 0.15}
\definecolor{mitblue}{rgb}{0.88,0.95,0.96}
\definecolor{lightblue}{rgb}{0.90, 0.95, 0.99}

\newcommand{\DiagDataset}{Diagnostic Dataset\xspace}

\definecolor{main}{HTML}{4472C4}    
\definecolor{sub}{HTML}{EBF4FF}     
\newtcolorbox{boxA}{
    enhanced, breakable,
    boxrule = 0pt,
    colback = sub,
    borderline west = {2pt}{0pt}{main}, 
    borderline east = {2pt}{0pt}{main}, 
}

\title{SciML Agents: Write the Solver, Not the Solution}

\author{
  Saarth Gaonkar$^*$\textsuperscript{1} \quad
  Xiang Zheng$^*$\textsuperscript{1} \quad
  Haocheng Xi\textsuperscript{1} \quad
  Rishabh Tiwari\textsuperscript{1} \quad \\ \AND
  Kurt Keutzer\textsuperscript{1} \quad
  Dmitriy Morozov\textsuperscript{2} \quad
  Michael W. Mahoney\textsuperscript{1,2,3} \quad
  Amir Gholami\textsuperscript{1,3} \\
  \\
  \textsuperscript{1}UC Berkeley \quad
  \textsuperscript{2}LBNL \quad
  \textsuperscript{3}ICSI
}

\begin{document}

\maketitle

\begin{abstract}
A large body of recent work in scientific machine learning (SciML) aims to tackle scientific tasks \emph{directly} by predicting target values with neural networks (e.g., physics-informed neural networks, neural ODEs, neural operators, etc.), but attaining high accuracy and robustness has been challenging. 
We explore an alternative view: use large language models (LLMs) to \emph{write code} that leverages decades of numerical algorithms.
This shifts the burden from learning a solution function to making domain-aware numerical choices. We ask whether LLMs can act as \emph{SciML agents} that, given a natural-language ODE description, generate runnable code that is \emph{scientifically appropriate}—selecting suitable solvers (stiff vs.\ non-stiff), setting tolerances, and enforcing stability checks.
There is currently no benchmark to measure this kind of capability for scientific computing tasks.
As such, we first introduce two new datasets: a diagnostic dataset of adversarial ``misleading'' problems; and a large-scale benchmark of 1,000 diverse ODE tasks. 
The diagnostic set contains problems whose superficial appearance suggests stiffness, and that require algebraic simplification to demonstrate non-stiffness; and the large-scale benchmark spans stiff and non-stiff ODE regimes.
We evaluate open- and closed-source LLM models along two axes: (i) unguided versus guided prompting with domain-specific knowledge; and (ii) off-the-shelf versus fine-tuned variants. 
Our evaluation measures both executability and numerical validity against reference solutions. 
We find that with sufficient context and guided prompts, newer instruction-following models achieve high accuracy on both criteria. In many cases, recent open-source systems (e.g., the Qwen3 family) perform strongly without fine-tuning, while older or smaller models still benefit from fine-tuning. Overall, our preliminary results indicate that careful prompting and fine-tuning can yield a specialized LLM agent capable of reliably solving simple ODE problems.
We have open-sourced the code and benchmark at \href{https://github.com/SqueezeAILab/sciml-agent}{\texttt{https://github.com/SqueezeAILab/sciml-agent}}.

\end{abstract}

\section{Introduction}

Recently, several approaches have used machine learning to solve scientific tasks by predicting solutions directly with neural networks.
Notable works include:
Physics-Informed Neural Networks~\cite{raissi2019physics}, 
neural ODEs~\cite{neuralODE18_TR,gholami2019anode}, neural operators~\cite{lu2021learning,bhattacharya2021model,kovachki2023neural,nelsen2021random,anandkumar2020neural,patel2021physics}, Finite-Dimensional operators~\cite{adler2017solving,bhatnagar2019prediction,khoo2021solving}, neural FEMs~\cite{yu2018deep,bar2019unsupervised}, Fourier Neural Operators~\cite{li2020fourier,fan2019bcr},
and neural discrete equilibria~\cite{benitez2025neural} to name a few.
Although these approaches have their strengths, training these models to output
the solution with high accuracy can be challenging~\cite{failure21_TR}.

By contrast, decades of numerical analysis offer robust algorithms for scientific computing problems. 
This raises a natural question: instead of predicting target values directly, why not use AI models to write code that leverages established numerical solvers?

This direction builds on the rapid progress of large language models (LLMs) in reasoning and code generation. General-purpose coding agents (e.g., IDE-integrated assistants) have excelled at code correctness, but not necessarily at domain-appropriate numerical choices. We therefore ask: given a natural-language description of a physical system, e.g., governed by an Ordinary Differential Equation (ODE)\cite{10.5555/153158}, can an LLM generate code that not only compiles correctly and runs, but is also \emph{scientifically appropriate}? If so, LLMs can serve as a bridge from problem descriptions to mature numerical solvers, yielding accurate solutions via principled computation.

While general-purpose coding assistants are becoming commonplace, scientific coding demands much more than syntactic correctness.
It requires domain-aware, numerically appropriate design choices. 
Examples include selecting the right solver for a stiff versus non-stiff problem, setting sensible error tolerances, and ensuring the stability of the solution. 
An incorrect choice can yield a scientifically invalid result, even if the code executes without error.

Prior work on general-purpose coding agents have largely focused on algorithmic correctness and passing unit tests. 
However, scientific computing tasks have nuanced requirements of numerical suitability and appropriate algorithmic choices that depend on the problem structure.
As such, a significant gap exists in the evaluation of LLMs for scientific code generation. 
In particular, there is currently no standard benchmark designed to systematically measure the scientific validity of LLM-generated code for solving ODEs. 
This leaves it unclear how robustly these models can reason about the underlying mathematical properties of a problem to select an optimal solution strategy.

In this work, we address this gap through the following contributions.
\begin{itemize}[leftmargin=*] 

    \item 
    \textbf{Robustness Evaluation of Symbolic Reasoning.}
    We design adversarial, \emph{misleading} ODEs whose superficial form suggests stiffness, but which algebraically simplifies to a non-stiff ODE. The task probes whether models perform genuine symbolic reduction rather than pattern matching, and whether they choose explicit methods when appropriate. We find that while older models struggle more with these examples, newer models demonstrate strong ``reasoning'' capabilities. In addition, we observe that providing the model with a guided prompt can significantly boost performance. See Section~\ref{subsec:misleading} for details.

    \item 
    \textbf{ODE-1000 Benchmark.}
    To evaluate scientific code generation capability of these models, we create the ODE-1000 benchmark, comprising 1,000 diverse ODE problems. We evaluate the performance by (i) checking if the generated code compiles correctly, and (ii) computing the resulting error of the final numerical solution against the ground truth. See Section ~\ref{subsec:ODE-1000 Dataset} for details.

    \item 
    \textbf{Prompting vs. Fine-Tuning.}
     Using the ODE-1000 dataset, 
     we conduct a detailed empirical analysis across a wide range of open- and closed-source LLMs.
     In particular, we evaluate the code generation quality by considering several scenarios: (i) zero-shot/bare minimum prompt, (ii) guided prompt, and (iii) fine-tuned models.
    We find that with careful guidance, LLMs can act as practical and reliable SciML agents for a wide range of simple ODE tasks. 
    Newer, larger models are particularly adept at this, often achieving high accuracy with prompting alone, while smaller models still benefit significantly from fine-tuning. 
    See \sect{results_sect} for details.
\end{itemize}

These preliminary results establish initial evidence that careful prompting and targeted fine-tuning can yield practical \emph{SciML coding agents} for simple ODE problems, and that they can provide a foundation for evaluating scientific code generation beyond syntactic correctness.

\section{Related Work}

\paragraph{LLM based Coding Agents.}
A variety of benchmarks have been introduced to evaluate LLMs’ coding capabilities. 
Classic datasets such as APPS~\cite{hendrycks2021apps}, HumanEval~\cite{chen2021codex}, and MBPP~\cite{austin2021mbpp} assess function-level code generation through unit tests, 
while DS-1000~\cite{lai2022ds1000} emphasizes data science code synthesis with Python libraries. 
More recent benchmarks such as SWE-bench~\cite{kocetkov2023swebench} capture real-world bug fixing in software repositories, ChronoLLM customize LLMs to generate PyChrono virtual experiments,
and CodeJudgeBench~\cite{mchen2025codejudgebench} investigates the reliability of LLMs as judges for code generation, repair, and test generation.

\paragraph{Coding Agents for Scientific Problems.}
LLMs have recently been applied to scientific computing, showing strong capabilities in symbolic reasoning and code generation. 
Benchmarks such as IneqMath~\cite{sheng2025ineqmath}, NeqLIPS~\cite{li2025lips},HypoGeniC~\cite{zhou2024hypothesis}, Mamo~\cite{huang2024mamo} and ASyMOB~\cite{shalyt2025asymob} underscore the expanding role of LLMs in mathematical problem solving; while systems like PAL~\cite{gao2022pal} and LOGIC-LM~\cite{pan2023logiclm} combine LLMs with interpreters or symbolic solvers to improve arithmetic and logical reasoning. 
In the scientific computing domain, frameworks like CodePDE~\cite{li2025codepde} have shown that LLMs can generate executable solvers for PDEs through reasoning, debugging, and test-time scaling, in some cases surpassing both human experts and specialized neural PDE solvers on benchmarks. Another work is PDE-Controller \cite{soroco2025pdecontroller} which leverages LLMs to 
allow control PDE-based tasks through natural language.

Another important work in this direction is~\cite{jiang2025deepseek} which recently benchmarked DeepSeek, ChatGPT, and Claude on a wide range of scientific computing and SciML tasks, including ODEs, PDEs, FEMs, and PINNs. 
Our work complements theirs: while~\cite{jiang2025deepseek} studies cross-model performance differences of closed-source models, we focus on studying the reasoning capability of 
open/closed-source models, and explore the impacts of
guided prompting vs finetuning of these models on
the coding capability of these models on a new
set of datasets that we introduce.

\section{Problem Setup}

In this section, we first describe the problem formulation and then the two benchmark datasets that are used to evaluate the scientific code generation capability of the LLMs.

\subsection{Task Definition}

We focus mainly on the following task: given a natural-language description of a physical or mathematical system governed by an ODE, how can we make an LLM generate a complete and executable Python program that can numerically solve the ODE?

A successful LLM output should (of course) return executable code, correct parameters, and a relatively low numerical error. 
However, the code should also be ``scientifically appropriate,'' e.g., meaning that the LLM should demonstrate an understanding of the underlying mathematical properties of the ODE to make informed design decisions. 
Here are examples of key factors to consider.
\begin{itemize}[leftmargin=*]
    \item \textbf{Solver Selection.} An appropriate numerical solver must be chosen (e.g., an explicit Runge-Kutta method for non-stiff problems versus an implicit method like BDF for stiff problems).
    \item \textbf{Code Execution.} Executable and correct Python code must be generated to solve the ODE problem, setting sensible parameters, such as time intervals and step sizes.
    \item \textbf{Numerical Validity.} The final solution produced by the code must achieve a low numerical error, when compared against a high-fidelity ground-truth or analytical solution.
\end{itemize}

\subsection{Benchmark datasets}

A significant challenge we faced in evaluating LLMs for scientific tasks is the lack of benchmarks that explicitly account for the numerical and mathematical nuances required in scientific computation. 
Existing code generation datasets primarily emphasize algorithmic correctness and passing unit tests, but they do not capture the scientific validity/accuracy of the final solution. This is particularly problematic because for scientific computing problems there are many syntactically valid arguments and solvers that could be used, but only relatively few will lead to accurate solutions.

To address this gap, we first introduce a benchmark suite composed of two novel datasets. 
First, we present a specialized \DiagDataset{} that contains problems that are intentionally ``misleading'' to probe the LLM's ability for deeper mathematical reasoning, rather than solely relying on superficial pattern matching. 
Second, we present a large-scale dataset of 1,000 diverse ODE problems derived from natural-language descriptions, designed to test the end-to-end task of generating complete and numerically accurate solver code.

\subsubsection{Symbolic-Equivalence  \DiagDataset{}}
\label{subsec:misleading}

In numerical ODE integration, \emph{explicit} solvers advance the state directly from the current value. 
They are simple and efficient, but they can become unstable on stiff systems. 
In contrast, \emph{implicit} solvers require solving for the unknown future state.
They are typically more stable (though costlier), making them preferable for stiff or constrained dynamics. 
In practice, scientific computing experts first assess whether a problem is stiff, which then strongly influences their solver choice.

Here, we test whether current LLMs can correctly classify ODEs as stiff or non-stiff, an essential step for choosing the \emph{right} algorithm when generating code, and whether they rely on superficial cues (e.g., large coefficients) or perform nontrivial symbolic reasoning.

\paragraph{Dataset Construction.} 
The  \DiagDataset{} comprises three subsets that appear \emph{stiff} before simplification, but that reduce to non-stiff forms: a trigonometric subset, an inverse function subset, and an algebraic subset, described as follows.

\begin{boxA}
\begin{enumerate}[leftmargin=1.5em, itemsep=2pt, topsep=2pt]
\item \textbf{Trigonometric subset} uses trigonometric identities, such as $\sin(t+\pi/2)+\cos t=0$, to simplify the equation.
\item \textbf{Inverse-function subset} uses inverse operators collapse, such as $\arcsin(\sin t)=t$ on $[-\pi/2,\pi/2]$, to simplify the equation.
\item \textbf{Algebraic subset} uses expansions/factorizations, such as $(x+1)^3 - 6(x^2 + 1) - (x-1)^3 = 0$,  that remove large coefficients after simplification.
\end{enumerate}
\end{boxA}
In each case, the coefficients appear large, but they can be simplified through mathematical identities. 
Consequently, the correct choice is always an explicit method. 
We report accuracy as the proportion of instances for which the model selects an explicit solver. This setup probes basic symbolic reduction capability, rather than more simple pattern matching, by the LLM.

\subsubsection{ODE-1000 Dataset}
\label{subsec:ODE-1000 Dataset}

\paragraph{Overview.} 
The ODE-1000 dataset is a curated dataset consisting of 1,000 ODE problems paired with executable Python solutions. 
The dataset is designed to evaluate the ability of LLMs to derive a correct ODE and problem setup from a natural-language description, and to produce syntactically valid, numerically stable solver code with an appropriate method choice (e.g., implicit vs. explicit solver). 
Below, we describe the generation pipeline, example structure, and coverage.

\paragraph{Generation.} 
We synthesize candidate problems with GPT 4.1 (prompt templates in Appendix~\ref{box:data_generation}) by letting the model (1) derive the ODE from a realistic scenario, where the description is \emph{natural language}, (2) specify an appropriate solver method and the reasoning trace behind the choice, and (3) produce the equation and the initial conditions in SymPy format, and the Python code to solve the equation using the solve\_ivp function in SciPy library. 
The prompt is designed to enforce \emph{diversity} across: stiff vs. non-stiff dynamics, orders, time intervals, magnitudes and signs of coefficients, and solver coverage. 
The target distribution is visualized in Figure~\ref{fig:ode_dataset_visualization} and \app{ode_statistics}.

\paragraph{Example structure.} 
We present an example of the created dataset for clarity. 
Each item is stored as JSON with five fields, as follows.

\begin{boxA}
\begin{itemize}
    \item \texttt{description}: A capacitor is initially charged to 10 volts and is discharging through a resistor. The voltage V(t) across the capacitor decreases at a rate proportional to the voltage itself, with a time constant of 2 seconds. Find V(t) for t from 0 to 10 seconds, using 200 evaluation points.
    \item \texttt{sympy\_eq}: \texttt{Eq(Derivative(y(t), t), -y(t)/2)}
    \item \texttt{initial\_condition}: \texttt{y(0): 10}
    \item \texttt{reasoning}: This is a classic first-order linear ODE for exponential decay: dV/dt = -V/2. The time constant is given as 2 seconds. The equation is not stiff, so RK45 is suitable. The time interval and number of points allow us to observe the voltage decay smoothly.
    \item \texttt{code}:

\definecolor{codeblue2}{rgb}{0,0,1}
\lstset{
  backgroundcolor=\color{white},
  basicstyle=\fontsize{7.2pt}{7.2pt}\ttfamily\selectfont,
  columns=fixed,
  breaklines=true,
  captionpos=b,
  keywordstyle=\fontsize{7.2pt}{7.2pt}\color{codeblue2},
    emph={mask},
    emphstyle={\color[RGB]{255,52,179}},
}
\begin{lstlisting}[language=python]
def f(t, V):
    return -V / 2

t_eval = np.linspace(0, 10, 200)
sol = solve_ivp(f, [0, 10], [10], t_eval=t_eval, method='RK45')
\end{lstlisting}

\end{itemize}
\end{boxA}

\paragraph{Verification and filtering.} 
After the dataset creation, for each candidate, we run the provided \texttt{code} in a sandbox to ensure it is syntactically valid, and \texttt{solve\_ivp} returns a solution on the supplied \texttt{t\_eval}. 
Candidates that fail to execute are discarded. 

We then obtain $y_{\text{analytical}}(t)$ using SymPy \texttt{dsolve} from \texttt{sympy\_eq} and \texttt{initial\_condition}. 
We evaluate both the numerical solution $y_{\text{numerical}}(t_i)$ and the analytical solution on the same grid and compute the relative $L_2$ error:
\[
\mathrm{Rel}\ L_2 = \frac{\lVert y_{\text{analytical}} - y_{\text{numerical}} \rVert_2}{\lVert y_{\text{analytical}} \rVert_2}.
\]
Examples with $\mathrm{Rel}\ L_2 > 0.01$ are rejected and filtered out, as they are numerically unstable and thus not suitable for benchmarking. 
After filtering, this process yields 1000 verified examples that emphasize realistic phrasing and solver sensitivity.

\section{SciML Agent Setup: Models, Prompting, and Fine-Tuning}

To evaluate the capabilities of LLMs as scientific computing agents for solving ODEs, we consider a diverse range of open/closed-source models, two distinct prompting strategies to analyze the impact of guidance, and a fine-tuning recipe to enhance the performance of small models.

\subsection{Models}

Our evaluation includes a wide array of both \textbf{open- and closed-source LLM families}, to ensure broad coverage of the current LLM landscape. 
Open-source models we tested include the Llama series (Llama2, Llama3, Llama3.1) and the Qwen series (Qwen1, Qwen1.5, Qwen2, Qwen2.5, Qwen3). For a closed-source model, we tested OpenAI's GPT-4.1.

We investigated a wide range of \textbf{model sizes}, from small models like Qwen3-0.6B to large-scale models such as Llama3.1-405B, to analyze the correlation between parameter count and scientific reasoning ability. 
We also consider model \textbf{vintage}, distinguishing between ``older'' models (released in 2023--2024) and ``newer'' instruction-following models (from 2025). 
This allows us to track the evolution of reasoning and instruction-following capabilities over successive generations.

\subsection{Prompting Methods}
\label{sec:prompting_regimes}
While pretrained language models possess strong symbolic capabilities, we observe that these abilities often remain underutilized in zero-shot settings, at least without a proper prompting structure. 
To explore this, we assess model performance under two different prompting methodologies to understand the effect of contextual guidance.

\begin{itemize}[leftmargin=*]
    \item \textbf{Unguided Prompt.} This prompt provides only minimal instructions. For the diagnostic tasks, it simply asks the model to classify an ODE and choose a solver. This regime is designed to test the model's out-of-the-box reasoning ability.

    \item \textbf{Guided Prompt.} In contrast, this prompt is engineered to provide substantial guidance. It explicitly instructs the model to perform step-by-step reasoning, simplify expressions algebraically, and analyze coefficient behavior before arriving at a solution. The prompt enriches the model's decision-making process by embedding domain-specific heuristics. 
\end{itemize}

Details on these two prompts can be found in Appendix~\ref{box:guided}.

\subsection{Fine-Tuning}

While newer, larger models often perform well with guided prompting, fine-tuning can be essential for improving the reliability of smaller models. We employ a supervised fine-tuning (SFT) approach to create specialized ODE-solving agents.

\begin{itemize}[leftmargin=*]
    \item \textbf{Training Data.} The training data is derived from our \textbf{ODE-1000 benchmark}. We use 900 examples for training and reserve the remaining 100 for testing. For the Qwen models, the fine-tuning data included the intermediate reasoning traces to better leverage their native reasoning capabilities, whereas for the Llama models, only the final Python code was used.

    \item \textbf{Settings.} We performed \textbf{full parameter fine-tuning} rather than a parameter-efficient method like LoRA. Models were trained for 3 epochs across 8 GPUs.
\end{itemize}

\section{Results}\lblsect{results_sect}

In this section, we summarize our main results. We first discuss the symbolic reasoning capability and how that has evolved with the newer models, and then we present the ODE-1000 benchmark~results.

\subsection{Symbolic-Equivalence \DiagDataset{} Evaluation}

\begin{figure}[t]
    \centering
    \includegraphics[width=0.95\linewidth]{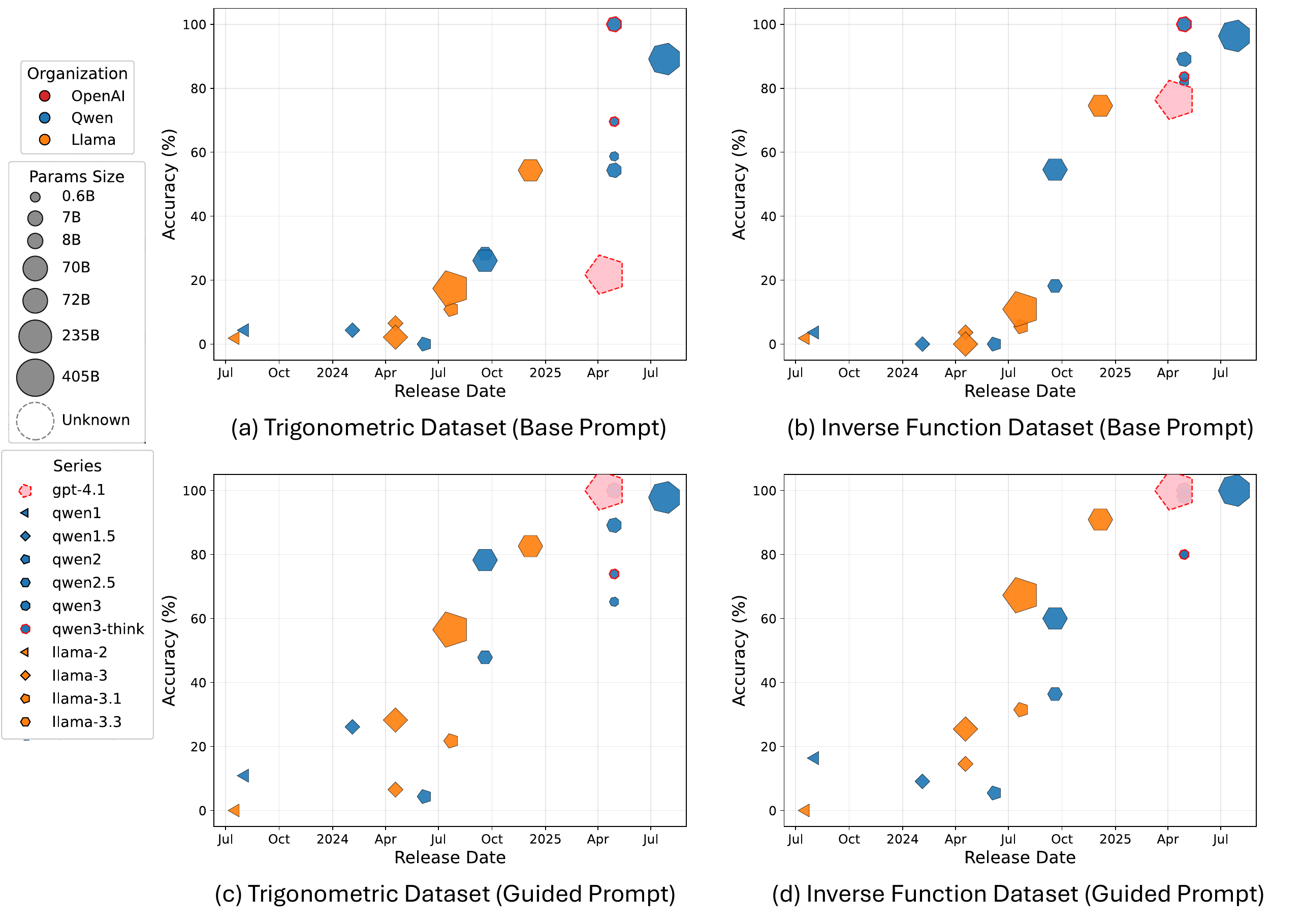}
    \caption{Evaluation of LLMs across varying release dates and model sizes. (a)–(b) present results on the trigonometric and inverse function tasks using the base prompt, while (c)–(d) present results on the same tasks using the guided prompt.
    We can see a clear trend where newer models are able to achieve higher accuracy. Furthermore, enabling thinking/reasoning mode can boost the accuracy by up to 25\%.
}
    \label{fig:mislead_model_eval_main}
\end{figure}

\paragraph{Older Model Performance on \DiagDataset{}}
We first evaluate models released in 2023 and 2024, including the LLaMA series (2, 3, 3.1, and 3.3) and Qwen series (1, 1.5, 2, and 2.5). 
Results are summarized in Table~\ref{tab:grouped_prompt_performance} and Figure~\ref{fig:mislead_model_eval_main}. 
We observe the following.
\begin{itemize}[leftmargin=*]
    \item 
    \textbf{Rapid improvement across LLaMA generations.} 
    The reasoning ability of LLaMA models improves significantly from LLaMA 3 to LLaMA 3.3. 
    \item 
    \textbf{Architecture matters more than size.} 
    Newer-generation models often outperform earlier ones, even when smaller in size. 
    For example, the smallest model in a newer generation frequently performs better than the largest model from the previous generation.
    \item 
    \textbf{Scale-performance correlation.} 
    Reasoning ability generally improves with model size.
\end{itemize}

\paragraph{Newer Model Performance on \DiagDataset{}}
We next evaluate models released in 2025, including the Qwen 3 series and GPT-4.1. For both Qwen3-8B and Qwen3-0.6B, we compare their reasoning ability with and without the ``thinking mode'' (where the model is prompted to reason step by step before producing a final answer). 
These results are summarized in Table~\ref{tab:grouped_prompt_performance}, and 
they lead to the following key observations. 
\begin{itemize}[leftmargin=*]
\item 
\textbf{Thinking mode significantly enhances reasoning performance.}
The thinking mode leads to notable improvements in accuracy, particularly on the algebraic subset. 
For instance, the performance of Qwen3-0.6B improves from 37.04\% to 75.93\% when using thinking mode. 
\item 
\textbf{Newer models exhibit stronger implicit reasoning.} 
Compared to earlier models, newer ones tend to engage in intermediate reasoning, rather than relying solely on pattern matching. As shown in \autoref{app:mislead_failure_case}, there are traces of failed reasoning in which the model incorrectly assumes that a superficially large coefficient directly implies stiffness, without further mathematical derivation. 
We observe that newer models are more likely to attempt to derive coefficients or to apply relevant transformations, whereas older models are more prone to being misled by superficial patterns.
\end{itemize}

\begin{table*}[t]
\centering
\begin{threeparttable}
\small
\setlength{\tabcolsep}{5pt}
\caption{Accuracy (\%) across three mathematical diagnostic datasets and prompt strategies. 
We can see that providing better instructions in the prompt (Guided) increases the model significantly. This is particularly true for newer models which have better instruction following capability, which means that fine-tuning may not necessarily be required if enough context is provided to these models to generate proper code.
}
\begin{tabular}{cc|cc|cc|cc}
\toprule
 & & \multicolumn{2}{c|}{\textbf{Trigonometric}} & \multicolumn{2}{c|}{\textbf{Algebraic}} & \multicolumn{2}{c}{\textbf{Inverse Function}} \\
\textbf{Model} & \textbf{Size} & Base & Guided & Base & Guided & Base & Guided \\
\midrule
\multicolumn{8}{c}{\textit{Llama and Qwen - Small models}} \\
\midrule
Llama2 & 7B & \textbf{1.85\%} & 0\% & 0\% & \textbf{5.56\%} & \textbf{1.85\%} & 0\% \\
Llama3 & 8B & 6.5\% & 6.5\% & 18.5\% & \textbf{31.5\%} & 3.6\% & \textbf{14.6\%} \\
Llama3.1 & 8B & 10.9\% & \textbf{21.7\%} & 27.8\% & \textbf{44.4\%} & 5.5\% & \textbf{31.5\%} \\
\midrule
Qwen1 & 7B & 4.35\%& \textbf{10.87\%} & \textbf{12.96\%} & 5.56\%& 3.64\% & \textbf{16.36\%} \\
Qwen1.5 & 7B & 4.35\%& \textbf{26.1\%} & 7.41\%& \textbf{25.93\%} & 0\% & \textbf{9.09\%} \\
Qwen2 & 7B & 0\%& \textbf{4.35\%} & 11.11\%& \textbf{14.81\%} & 0\%&\textbf{5.45\%}\\
Qwen2.5 & 7B & 28.3\% & \textbf{47.8\%} & 61.1\% & \textbf{77.8\%} & 18.2\% & \textbf{36.4\%} \\
\midrule
\multicolumn{8}{c}{\textit{Llama and Qwen - Big models}} \\
\midrule
Llama3 & 70B & 2.2\% & \textbf{28.3\%} & 38.9\% & \textbf{87.0\%} & 0.0\% & \textbf{25.5\%} \\
Qwen2.5 & 72B & 26.1\% & \textbf{78.3\%} & 53.7\% & \textbf{83.3\%} & 54.5\% & \textbf{60.0\%} \\
Llama3.3 & 70B & 54.4\% & \textbf{82.6\%} & 81.5\% & \textbf{92.6\%} & 74.5\% & \textbf{90.9\%} \\
\midrule
Llama3.1 & 405B & 17.4\% & \textbf{56.5\%} & 50.0\% & \textbf{83.3\%} & 10.9\% & \textbf{67.3\%} \\
Qwen3 & 235B & 89.1\% & \textbf{97.8\%} & 100.0\% & \textbf{100.0\%} & 96.4\% & \textbf{100.0\%} \\
GPT-4.1 & - & 21.7\% & \textbf{100.0\%} & 92.6\% & \textbf{100.0\%} & 76.4\% & \textbf{100.0\%} \\
\midrule
\multicolumn{8}{c}{\textit{Reasoning Models}} \\
\midrule
Qwen3 & 0.6B & 58.7\% & \textbf{65.2\%} & 37.0\% & \textbf{51.9\%} & \textbf{81.8\%} & 80.0\% \\
Qwen3$^\dagger$ & 0.6B & 69.6\% & \textbf{73.9\%} & 75.9\% & \textbf{79.6\%} & \textbf{83.6\%} & 80.0\% \\
Qwen3 & 8B & 54.4\% & \textbf{89.1\%} & 77.8\% & \textbf{79.6\%} & 89.1\% & \textbf{98.2\%} \\
Qwen3$^\dagger$ & 8B & 100.0\% & 100.0\% & 100.0\% & 100.0\% & 100.0\% & 100.0\% \\
\bottomrule
\end{tabular}
\begin{tablenotes}
    \item[$\dagger$] Activate reasoning mode
\end{tablenotes}
\label{tab:grouped_prompt_performance}
\end{threeparttable}
\end{table*}

\subsubsection{The Role of Prompting on the \DiagDataset{}}
 
We analyze model performance using a \textit{base prompt} and a \textit{guided prompt} (recall Section~\ref{sec:prompting_regimes}). 
We observe that when using the base prompt, the model exhibits limited performance on the \DiagDataset{}. 
As shown in Table~\ref{tab:grouped_prompt_performance}, by switching to a guided prompt that explicitly encourages step-by-step reasoning and expression simplification, we observe a greatly improved performance and reveal the following observations. 
\begin{itemize}[leftmargin=*]
    \item 
    \textbf{Larger models demonstrate stronger instruction-following capabilities.}
    Within the same model family, we observe that models with more parameters consistently achieve better performance. 
    This suggests that scaling up model size enhances their ability to follow prompts effectively.
    \item 
    \textbf{Mathematical simplification abilities have improved significantly in recent models.}
    Compared to earlier generations, newer models such as the Qwen3 series and GPT-4.1 achieve over 80\% accuracy—even for relatively small models like Qwen3-0.6B.
    This indicates substantial progress in the models' capacity for step-by-step reasoning and symbolic manipulation.
\end{itemize}

Importantly, model failure is \emph{not} simply due to insufficient knowledge or representational capacity, but rather due to the absence of explicit guidance and structured reasoning incentives. 
As shown in \app{prompts_of_diagnostic}, the guided prompt adds an explicit instruction, guiding the model toward multi-step reasoning and suppressing pattern-matching behaviors.

\subsection{ODE Dataset Evaluation}
\subsubsection{Evaluation Criteria}
We test model performance on the ODE-1000 dataset using three main metrics.
\begin{itemize}[leftmargin=*]
    \item
    \textbf{Code execution rate.} We measure the proportion of times the code successfully compiles and runs end-to-end in a sandbox.
    \item 
    \textbf{Accuracy.} Among the examples in which the code runs successfully, the accuracy is defined as the fraction of instances in which the relative error $L_2$ with respect to the ground truth (SymPy closed-form solution) is less than 0.01.
    \item 
    \textbf{Mean relative $L_2$ error.} We also calculate the mean relative error of the code outputted by the model with respect to the ground truth, where we clamp to 1 to alleviate the effect of extreme outliers.
\end{itemize}

\subsubsection{ODE Dataset Results}

We observe that fine-tuning produces quantitative improvements for small models. 
See Table~\ref{tab:fine-tuning} for a summary.
For example, for the Qwen-3-0.6B model, the code execution rate jumps from $27\%$ to $97\%$, the mean relative $L_2$ drops from $0.3705$ to $0.1043$, and the accuracy among successful runs increases from $62.96\%$ to $87.63\%$. Llama2-7B shows a similar improvement, where execution goes from $21\%\!\rightarrow\!100\%$ and accuracy $47.62\%\!\rightarrow\!87\%$. 

These improvements can be mainly classified into two categories.
\begin{itemize}[leftmargin=*]
    \item 
    \textbf{Fewer syntax errors.} 
    Fine-tuning the models with the ODE-1000 dataset can greatly reduce syntax errors, which is necessary for ODE solvers to produce executable code.
    \item 
    \textbf{Improved numerical validity.} 
    Fine-tuning also greatly improves numerical accuracy by enhancing the model's ability to correctly derive and set up an equation and choose the correct solvers.
\end{itemize}

For both of these reasons, our results demonstrate that, for smaller and/or older models, fine-tuning can be an effective way to improve performance. 
On the other hand, for newer reasoning models that are sufficiently large (such as Qwen3-8B), fine-tuning is often not necessary; instead, careful prompting with in-context examples (evaluation prompt in Appendix~\ref{box:evaluation_prompt}) is enough to achieve near 100\% performance.%
\footnote{It is important to note that further fine-tuning may be also required if the Agent has to generate code 
in completely new domains, on which the base model has not been extensively trained. However, the scope
that we considered here involves common ODEs for which pretrained models have some general knowledge.}

\begin{table}[t]
\centering
\caption{Performance of various models on the ODE dataset. We can clearly see that fine-tuning helps, particularly for older or smaller models. However, as we saw in Table~\ref{tab:grouped_prompt_performance}, newer models can achieve high accuracy without finetuning if given enough context and instructions.}
\label{tab:fine-tuning}
    \begin{tabular}{l|ccc}
    \toprule
         Model\phantom{XXXXXXX} &  Code Execution Rate&  Accuracy& Relative L2 Error \\
    \midrule
         Qwen2-7B &  93\% &  91.40\% & 0.0841\\
         \hfill \textit{+ Finetuning} &  99\%&  94.95\% & 0.0459\\
         Qwen3-0.6B &  27\% &  62.96\% & 0.3705\\
         \hfill w/ reasoning &  27\% &  70.37\% & 0.2913\\
         \hfill \textit{+ Finetuning} &  97\% &  87.63\% & 0.1043\\
         Llama-7B & N/A & N/A & N/A \\
         \hfill \textit{+ Finetuning} & 100\% & 87\% &0.1261\\
         Llama2-7B& 21\%& 47.62\% &0.4645\\
         \hfill \textit{+ Finetuning} & 100\% & 87\% & 0.1169\\
         Llama3-8B&  96\%&  89.58\% & 0.0734\\
         \hfill \textit{+ Finetuning} & 98\% & 94.9\% & 0.0516\\ \bottomrule
    \end{tabular}
    \begin{tablenotes}
        \centering
        \footnotesize
        \item N/A means the model does not have instruction following ability.
    \end{tablenotes}
\end{table}

\section{Conclusion and Future Work}

We asked whether LLMs can serve as \emph{SciML agents} that translate natural-language ODE descriptions into runnable, \emph{scientifically appropriate} code. To study this, we introduced two benchmarks: a Symbolic-Equivalence diagnostic suite that probes algebraic simplification and solver choice, and \textsc{ODE-1000}, a large-scale collection of diverse problems. Our evaluation, measuring both \emph{executability} and \emph{numerical validity} against reference solutions, shows that careful guidance substantially improves performance: newer instruction-following models often achieve high accuracy with domain-aware prompting, while smaller or older models benefit markedly from fine-tuning. Together, these results indicate a practical path to specialized LLM agents that leverage mature numerical solvers rather than predicting solutions directly, and they establish a foundation for assessing scientific code generation beyond syntactic correctness.

\paragraph{Limitations and future work.}
Our study targets simple ODE settings; extending to richer regimes is an immediate next step. We plan to broaden the benchmarks to include systems with event handling and time-varying stiffness, multi-dimensional and higher-order dynamics, chaotic and stiff regimes with sharper diagnostics, boundary-value and parameter-estimation tasks, and ultimately PDEs. Methodologically, we aim to strengthen agent tooling (e.g., symbolic simplification and stability checks), incorporate property-based tests and invariants, and examine cross-library generalization. These directions will help stress-test scientific appropriateness, reliability, and robustness at scale.

\begin{ack}
We would like to thank Aditya Tomar, Amin Tutoonforoush, and Arnur Nigmetov for helpful feedback.
We also acknowledge gracious support from the FuriosaAI, Intel, Apple, NVIDIA, Macronix, and Mozilla team.
We also appreciate the support from Microsoft through their Accelerating Foundation Model Research.
Prof. Keutzer's lab is sponsored by the Intel corporation, UC Berkeley oneAPI Center of Excellence, Intel VLAB team, as well as funding through BDD and BAIR.
MWM also acknowledges DARPA, NSF, the DOE Competitive Portfolios grant, and the DOE SciGPT grant.

\end{ack}

{
    \small
    \bibliographystyle{unsrt}
    \bibliography{neurips_2025}

\begin{thebibliography}{10}

\bibitem{raissi2019physics}
Maziar Raissi, Paris Perdikaris, and George~E. Karniadakis.
\newblock Physics-informed neural networks: A deep learning framework for
  solving forward and inverse problems involving nonlinear partial differential
  equations.
\newblock {\em Journal of Computational Physics}, 378:686--707, 2019.

\bibitem{neuralODE18_TR}
Ricky T.~Q. Chen, Yulia Rubanova, Jesse Bettencourt, and David Duvenaud.
\newblock Neural ordinary differential equations.
\newblock In {\em Advances in Neural Information Processing Systems (NeurIPS)},
  volume~31, 2018.

\bibitem{gholami2019anode}
Amir Gholami, Kurt Keutzer, and George Biros.
\newblock Anode: Unconditionally accurate memory-efficient gradients for neural
  odes.
\newblock In {\em Proceedings of the 28th International Joint Conference on
  Artificial Intelligence (IJCAI)}, pages 730--736, 2019.

\bibitem{lu2021learning}
Lu~Lu, Pengzhan Jin, Guofei Pang, Zhongqiang Zhang, and George~E. Karniadakis.
\newblock Learning nonlinear operators via deeponet based on the universal
  approximation theorem of operators.
\newblock {\em Nature Machine Intelligence}, 3(3):218--229, 2021.

\bibitem{bhattacharya2021model}
Kaushik Bhattacharya, Bamdad Hosseini, Nikola~B Kovachki, and Andrew~M Stuart.
\newblock Model reduction and neural networks for parametric pdes.
\newblock {\em The SMAI journal of computational mathematics}, 7:121--157,
  2021.

\bibitem{kovachki2023neural}
Nikola~B. Kovachki, Zongyi Li, Burigede Liu, Kamyar Azizzadenesheli, Kaushik
  Bhattacharya, Andrew~M. Stuart, and Anima Anandkumar.
\newblock Neural operator: Learning maps between function spaces with
  applications to pdes.
\newblock {\em Journal of Machine Learning Research}, 24(89):1--97, 2023.

\bibitem{nelsen2021random}
Nicholas~H Nelsen and Andrew~M Stuart.
\newblock The random feature model for input-output maps between banach spaces.
\newblock {\em SIAM Journal on Scientific Computing}, 43(5):A3212--A3243, 2021.

\bibitem{anandkumar2020neural}
Anima Anandkumar, Kamyar Azizzadenesheli, Kaushik Bhattacharya, Nikola
  Kovachki, Zongyi Li, Burigede Liu, and Andrew Stuart.
\newblock Neural operator: Graph kernel network for partial differential
  equations.
\newblock In {\em ICLR 2020 workshop on integration of deep neural models and
  differential equations}, 2020.

\bibitem{patel2021physics}
Ravi~G Patel, Nathaniel~A Trask, Mitchell~A Wood, and Eric~C Cyr.
\newblock A physics-informed operator regression framework for extracting
  data-driven continuum models.
\newblock {\em Computer Methods in Applied Mechanics and Engineering},
  373:113500, 2021.

\bibitem{adler2017solving}
Jonas Adler and Ozan {\"O}ktem.
\newblock Solving ill-posed inverse problems using iterative deep neural
  networks.
\newblock {\em Inverse Problems}, 33(12):124007, 2017.

\bibitem{bhatnagar2019prediction}
Saakaar Bhatnagar, Yaser Afshar, Shaowu Pan, Karthik Duraisamy, and Shailendra
  Kaushik.
\newblock Prediction of aerodynamic flow fields using convolutional neural
  networks.
\newblock {\em Computational Mechanics}, 64(2):525--545, 2019.

\bibitem{khoo2021solving}
Yuehaw Khoo, Jianfeng Lu, and Lexing Ying.
\newblock Solving parametric pde problems with artificial neural networks.
\newblock {\em European Journal of Applied Mathematics}, 32(3):421--435, 2021.

\bibitem{yu2018deep}
Bing Yu et~al.
\newblock The deep ritz method: a deep learning-based numerical algorithm for
  solving variational problems.
\newblock {\em Communications in Mathematics and Statistics}, 6(1):1--12, 2018.

\bibitem{bar2019unsupervised}
Leah Bar and Nir Sochen.
\newblock Unsupervised deep learning algorithm for pde-based forward and
  inverse problems.
\newblock {\em arXiv preprint arXiv:1904.05417}, 2019.

\bibitem{li2020fourier}
Zongyi Li, Nikola Kovachki, Kamyar Azizzadenesheli, Burigede Liu, Kaushik
  Bhattacharya, Andrew Stuart, and Anima Anandkumar.
\newblock Fourier neural operator for parametric partial differential
  equations.
\newblock In {\em International Conference on Learning Representations (ICLR)},
  2021.

\bibitem{fan2019bcr}
Yuwei Fan, Cindy~Orozco Bohorquez, and Lexing Ying.
\newblock Bcr-net: A neural network based on the nonstandard wavelet form.
\newblock {\em Journal of Computational Physics}, 384:1--15, 2019.

\bibitem{benitez2025neural}
Jose Antonio~Lara Benitez, Junyi Guo, Kareem Hegazy, Ivan Dokmanic, Michael~W.
  Mahoney, and Maarten~V. de~Hoop.
\newblock Neural equilibria for long-term prediction of nonlinear conservation
  laws.
\newblock Technical report, arXiv preprint arXiv:2501.06933, 2025.

\bibitem{failure21_TR}
Aditi~S. Krishnapriyan, Amir Gholami, Shandian Zhe, Robert~M. Kirby, and
  Michael~W. Mahoney.
\newblock Characterizing possible failure modes in physics-informed neural
  networks.
\newblock In {\em Advances in Neural Information Processing Systems (NeurIPS)},
  volume~34, 2021.

\bibitem{10.5555/153158}
E.~Hairer, S.~P. N\o{}rsett, and G.~Wanner.
\newblock {\em Solving ordinary differential equations I (2nd revised. ed.):
  nonstiff problems}.
\newblock Springer-Verlag, Berlin, Heidelberg, 1993.

\bibitem{hendrycks2021apps}
Dan Hendrycks, Steven Basart, Saurav Kadavath, Mantas Mazeika, Akul Arora,
  Ethan Guo, Collin Burns, Samir Puranik, Horace He, Dawn Song, and Jacob
  Steinhardt.
\newblock Measuring coding challenge competence with apps.
\newblock In {\em NeurIPS 2021 Track on Datasets and Benchmarks}, 2021.

\bibitem{chen2021codex}
Mark Chen, Jerry Tworek, Heewoo Jun, Qiming Yuan, Henrique Ponde de~Oliveira
  Pinto, Jared Kaplan, Harri Edwards, Yuri Burda, Nicholas Joseph, Greg
  Brockman, Alex Ray, Raul Puri, Gretchen Krueger, Michael Petrov, Heidy
  Khlaaf, Girish Sastry, Pamela Mishkin, Brooke Chan, Scott Gray, Nick Ryder,
  Mikhail Pavlov, Alethea Power, Łukasz Kaiser, Mohammad Bavarian, Clemens
  Winter, Philippe Tillet, Felipe~Petroski Such, Dave Cummings, Matthias
  Plappert, Fotios Chantzis, Elizabeth Barnes, Ariel Herbert-Voss,
  William~Hebgen Guss, Alex Nichol, Alex Paino, Nikolas Tezak, Jie Tang, Igor
  Babuschkin, Suchir Balaji, Shantanu Jain, William Saunders, Christopher
  Hesse, Andrew~N. Carr, Jan Leike, Josh Achiam, Vedant Misra, Evan Morikawa,
  Alec Radford, Matthew Knight, Miles Brundage, Mira Murati, Katie Mayer, Peter
  Welinder, Bob McGrew, Dario Amodei, Sam McCandlish, Ilya Sutskever, and
  Wojciech Zaremba.
\newblock Evaluating large language models trained on code.
\newblock {\em arXiv preprint arXiv:2107.03374}, 2021.

\bibitem{austin2021mbpp}
Jacob Austin, Augustus Odena, Maxwell Nye, Maarten Bosma, Henryk Michalewski,
  David Dohan, Ellen Jiang, Carrie Cai, Michael Terry, Quoc~V. Le, and Charles
  Sutton.
\newblock Program synthesis with large language models.
\newblock {\em arXiv preprint arXiv:2108.07732}, 2021.

\bibitem{lai2022ds1000}
Yuhang Lai, Chengxi Li, Yiming Wang, Tianyi Zhang, Ruiqi Zhong, Luke
  Zettlemoyer, Wen-Tau Yih, Daniel Fried, Sida Wang, and Tao Yu.
\newblock Ds-1000: A natural and reliable benchmark for data science code
  generation.
\newblock In Andreas Krause, Emma Brunskill, Kyunghyun Cho, Barbara Engelhardt,
  Sivan Sabato, and Jonathan Scarlett, editors, {\em Proceedings of the 40th
  International Conference on Machine Learning (ICML)}, volume 202 of {\em
  Proceedings of Machine Learning Research}, pages 18319--18345. PMLR, 23--29
  Jul 2023.

\bibitem{kocetkov2023swebench}
Carlos~E. Jimenez, John Yang, Alexander Wettig, Shunyu Yao, Kexin Pei, Ofir
  Press, and Karthik~R. Narasimhan.
\newblock Swe-bench: Can language models resolve real-world github issues?
\newblock In {\em International Conference on Learning Representations (ICLR)},
  2024.
\newblock Oral.

\bibitem{mchen2025codejudgebench}
Hongchao Jiang, Yiming Chen, Yushi Cao, Hung-yi Lee, and Robby~T. Tan.
\newblock Codejudgebench: Benchmarking llm-as-a-judge for coding tasks.
\newblock {\em arXiv preprint arXiv:2507.10535}, 2025.

\bibitem{sheng2025ineqmath}
Jiayi Sheng, Luna Lyu, Jikai Jin, Tony Xia, Alex Gu, James Zou, and Pan Lu.
\newblock Solving inequality proofs with large language models.
\newblock {\em arXiv preprint arXiv:2506.07927}, 2025.

\bibitem{li2025lips}
Zenan Li, Zhaoyu Li, Wen Tang, Xian Zhang, Yuan Yao, Xujie Si, Fan Yang, Kaiyu
  Yang, and Xiaoxing Ma.
\newblock Proving olympiad inequalities by synergizing llms and symbolic
  reasoning.
\newblock In {\em International Conference on Learning Representations (ICLR)},
  2025.

\bibitem{zhou2024hypothesis}
Yangqiaoyu Zhou, Haokun Liu, Tejes Srivastava, Hongyuan Mei, and Chenhao Tan.
\newblock Hypothesis generation with large language models.
\newblock {\em arXiv preprint arXiv:2404.04326}, 2024.

\bibitem{huang2024mamo}
Xuhan Huang, Qingning Shen, Yan Hu, Anningzhe Gao, and Benyou Wang.
\newblock Mamo: A mathematical modeling benchmark with solvers.
\newblock {\em arXiv preprint arXiv:2405.13144}, 2024.

\bibitem{shalyt2025asymob}
Michael Shalyt, Rotem Elimelech, and Ido Kaminer.
\newblock Asymob: Algebraic symbolic mathematical operations benchmark.
\newblock {\em arXiv preprint arXiv:2505.23851}, 2025.

\bibitem{gao2022pal}
Luyu Gao, Aman Madaan, Shuyan Zhou, Uri Alon, Pengfei Liu, Yiming Yang, Jamie
  Callan, and Graham Neubig.
\newblock {PAL}: Program-aided language models.
\newblock In {\em Proceedings of the 40th International Conference on Machine
  Learning (ICML)}, volume 202 of {\em Proceedings of Machine Learning
  Research}, pages 10764--10799. PMLR, 2023.

\bibitem{pan2023logiclm}
Liangming Pan, Alon Albalak, Xinyi Wang, and William Wang.
\newblock Logic-{LM}: Empowering large language models with symbolic solvers
  for faithful logical reasoning.
\newblock In {\em Findings of the Association for Computational Linguistics:
  EMNLP 2023}, pages 3806--3824, Singapore, December 2023. Association for
  Computational Linguistics.

\bibitem{li2025codepde}
Shanda Li, Tanya Marwah, Junhong Shen, Weiwei Sun, Andrej Risteski, Yiming
  Yang, and Ameet Talwalkar.
\newblock Codepde: An inference framework for llm-driven pde solver generation.
\newblock {\em arXiv preprint arXiv:2505.08783}, 2025.

\bibitem{soroco2025pdecontroller}
Mauricio Soroco, Jialin Song, Mengzhou Xia, Kye Emond, Weiran Sun, and Wuyang
  Chen.
\newblock Pde-controller: {LLMs} for autoformalization and reasoning of {PDEs}.
\newblock In {\em Proceedings of the 42nd International Conference on Machine
  Learning}, volume 267 of {\em Proceedings of Machine Learning Research},
  Vancouver, Canada, 2025. PMLR.
\newblock ICML 2025.

\bibitem{jiang2025deepseek}
Qile Jiang, Zhiwei Gao, and George~Em Karniadakis.
\newblock Deepseek vs. chatgpt vs. claude: A comparative study for scientific
  computing and scientific machine learning tasks.
\newblock {\em Theoretical and Applied Mechanics Letters}, 15(3):100583, 2025.

\end{thebibliography}
}

\newpage
\counterwithin{figure}{section}
\counterwithin{table}{section}
\appendix


\section{Prompt Engineering for ODE-1000 Dataset}

In this section, we present prompts we adopt to generate and evaluate the ODE-1000 dataset.

\begin{tcolorbox}[float, floatplacement=h,colback=gray!5!white, colframe=gray!75!black, title=ODE Dataset Evaluation Prompt, label={box:evaluation_prompt}]
You are an expert in solving ODEs using Python. Your job is to generate code to solve an ODE problem by making a call to solve\_ivp with the correct parameters: correct function code, correct method (RK45, RK23, BDF, Radau, LSODA, DOP853, etc.), correct t\_eval, correct number of evaluation points, etc.
\\
\\
Here's an example problem: Population P(t) grows logistically according to dP/dt = 0.5*P*(1 - P/1000) with initial population 50. Find P(t) from t = 0 to t = 40 using 250 evaluation points.
\\
\\
Here's the corresponding example solution: def f(t, P):    return 0.5 * P * (1 - P / 1000) t\_eval = np.linspace(0, 40, 250) sol = solve\_ivp(f, [0, 40], [50], t\_eval=t\_eval, method='RK45')
\\
\\
Now, here's the problem you need to solve: {example["description"]}
\\
In the final output, only return the correct code (function and call to solve\_ivp), nothing else. Use "sol" to refer to the solution. Make sure that the code is syntactically correct, and do not output any other text.
\end{tcolorbox}

\begin{tcolorbox}[float, floatplacement=h,colback=gray!5!white, colframe=gray!75!black, title=ODE Data Generation Prompt, label={box:data_generation}]
Generate {batch\_size} diverse examples of ODEs that can be solved using scipy's solve\_ivp. Each example should include:
\\
1. A clear description of the ODE in terms of variables of your choice (you should always give away t\_eval and number of evaluation points)
\\
2. The correct corresponding sympy equation 
\\
3. The correct initial condition (in valid sympy format) 
\\
4. Reasoning to set up/solve the ODE with the correct solve\_ivp parameters
\\
5. The correct Python code using solve\_ivp with appropriate parameters
\\
\\

The sympy equation and initial conditions should be in terms of y(t) even if the original equation is given in terms of different variables.
\\
\\
The examples should be diverse in terms of:
\\
- Different types of ODEs (different orders, stiff, non-stiff, etc.)
\\
- Different correct usages of numerical methods (BDF, Radau, LSODA
for stiff problems and RK45, RK23, DOP853 for non-stiff problems)
\\
- Different time intervals
\\
- Different initial conditions
\\
\\

IMPORTANT: 
\\
-Generate real-world problems. The ODE should not be given and needs to be derived from the description. Make it as hard as possible; it should be very challenging to correctly derive the ODE, requiring substantial reasoning.
\\
-In the problem description, do not give away if the ODE is stiff or not stiff.
\\
-Generate an equal number of stiff and non-stiff problems.  
\\
-When generating a problem with a given method, make sure that the the problem results in a high error or solver instability if solved using a different method.
\\
-Ensure each example has a known closed-form solution that can be found.
\\
\\

Format the output as a JSON array with each example having only the following fields: description, sympy\_eqs, initial\_conditions, reasoning, code. 
\\
Here is an example:
\\
\{
\\
    "description": "A population P(t) grows at a rate proportional to its current size with a proportionality constant 0.3. The initial population is 100. Find P(t) from t = 0 to t = 20 using 200 points.",
    \\
    "sympy\_eq": "Eq(Derivative(y(t), t), 0.3 * y(t))",
    \\
    "initial\_condition": "{y(0): 100}",
    \\
    "reasoning": "The ODE is directly given: dP/dt = 0.3*P. This is a classic exponential growth equation, non-stiff and first order. RK45 is appropriate. The time domain [0, 20] and 200 evaluation points will capture the exponential curve well. The initial condition is mapped directly.",
    \\
    "code": "import numpy as np from scipy.integrate import solve\_ivp def f(t, y):    return 0.3 * y t\_eval = np.linspace(0, 20, 200) sol = solve\_ivp(f, [0, 20], [100], t\_eval=t\_eval, method='RK45')"
    \\
\}
\\
\\
Make sure the examples are challenging and diverse. Return ONLY the JSON array, with no additional text before or after. It must be a valid JSON array.
\end{tcolorbox}

\clearpage

\section{ODE Dataset Statistics}\lblapp{ode_statistics}

In this section, we present the distribution and statistics for solver methods and ODE orders for the ODE-1000 dataset.

\begin{figure}[h]
    \centering
    \includegraphics[width=\linewidth]{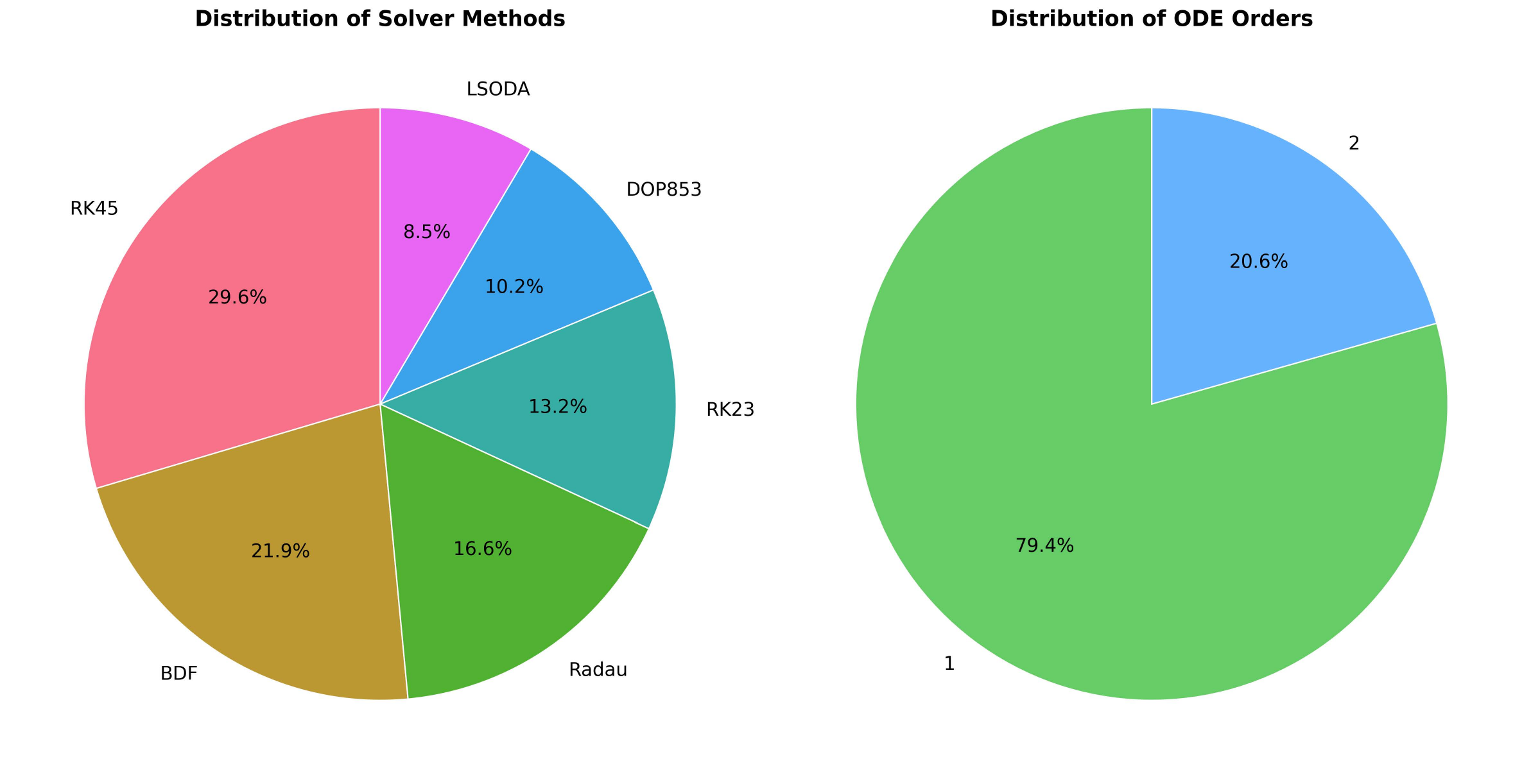}
    \caption{ODE-1000 dataset distribution of solver methods and ODE orders (1st/2nd).}
    \label{fig:ode_dataset_visualization}
\end{figure}

\clearpage


\section{Prompt Engineering for \DiagDataset{}}\lblapp{prompts_of_diagnostic}

In this section, we discuss prompts for the \DiagDataset{}.

The guided prompt introduces two key forms of inductive structure:
\begin{itemize}
    \item It embeds domain-specific heuristics for stiffness (e.g., "eigenvalues $\ll -1000$ imply stiffness"), giving the model actionable thresholds for decision-making.
    \item It enforces algebraic and symbolic simplification of the coefficients.
\end{itemize}

These differences are highlighted in red.

\begin{tcolorbox}[float, floatplacement=h,colback=gray!5!white, colframe=gray!75!black, title=Base System Prompt, label={box:base}]
You are an expert in numerical methods for solving Ordinary Differential Equations (ODEs).
    Your task is to determine whether an explicit or implicit solver should be used for the given ODE.
    Consider the following criteria: \\
    - Explicit solvers (like RK45, RK23, DOP853) are suitable for non-stiff ODEs \\
    - Implicit solvers (like BDF, Radau) are suitable for stiff ODEs \\
    - LSODA automatically switches between stiff and non-stiff solvers, making it ideal for ODEs with varying stiffness. \\
    You will be given an ODE \texttt{ode\_example} and corresponding \texttt{time\_interval}. \\
    Please analyze the ODE step by step: \\
    1. First, provide your detailed reasoning about why this ODE is stiff or non-stiff. \\
    2. Then, provide your final answer as either "explicit solver" or "implicit solver". Don't mention anything else, just the solver choice. \\
    Please format your response in json format as follows. IT MUST BE VALID JSON:  \\
    \{ \\
        "reason": "Your detailed analysis here", \\
        "answer": "explicit solver/implicit solver" \\
    \} \\
\end{tcolorbox}

\begin{tcolorbox}[float, floatplacement=h,colback=gray!5!white, colframe=gray!75!black, title=Guiding System Prompt, label={box:guided}]
 You are an expert in numerical methods for solving Ordinary Differential Equations (ODEs).
    Your task is to determine whether an explicit or implicit solver should be used for the given ODE.
    Consider the following criteria: \\
    - Explicit solvers (like RK45, RK23, DOP853) are suitable for non-stiff ODEs \\
    - Implicit solvers (like BDF, Radau) are suitable for stiff ODEs \\
    - LSODA automatically switches between stiff and non-stiff solvers, making it ideal for ODEs with varying stiffness. \\
    - \textcolor{red}{Stiff ODEs typically have large negative eigenvalues (much less than -1000)}
    - \textcolor{red}{Non-stiff ODEs have eigenvalues greater than -100} \\
    You will be given an ODE \texttt{ode\_example} and corresponding \texttt{time\_interval}. \\
    Please analyze the ODE step by step: \\
    \textcolor{red}{1. Given that the ODE is in the form of dy/dt = p(t)y + q(t), please first try to simplify the p(t) and q(t) to make it easier to solve.} \\
    2. Then, provide your detailed reasoning about why this ODE is stiff or non-stiff. \\
    3. Finally, provide your final answer as either "explicit solver" or "implicit solver". Don't mention anything else, just the solver choice. \\
    Please format your response in json format as follows. IT MUST BE VALID JSON: \\
    \{ \\
        "reason": "Your detailed analysis here", \\
        "answer": "explicit solver/implicit solver" \\
    \} \\
\end{tcolorbox}

\clearpage

\section{Model Evaluation on Diagnostic Dataset}\lblapp{mislead_model_eval_full}

In this section, we visualize the full result on the diagnostic dataset.

\begin{figure}[h]
    \centering
    \includegraphics[width=0.95\linewidth]{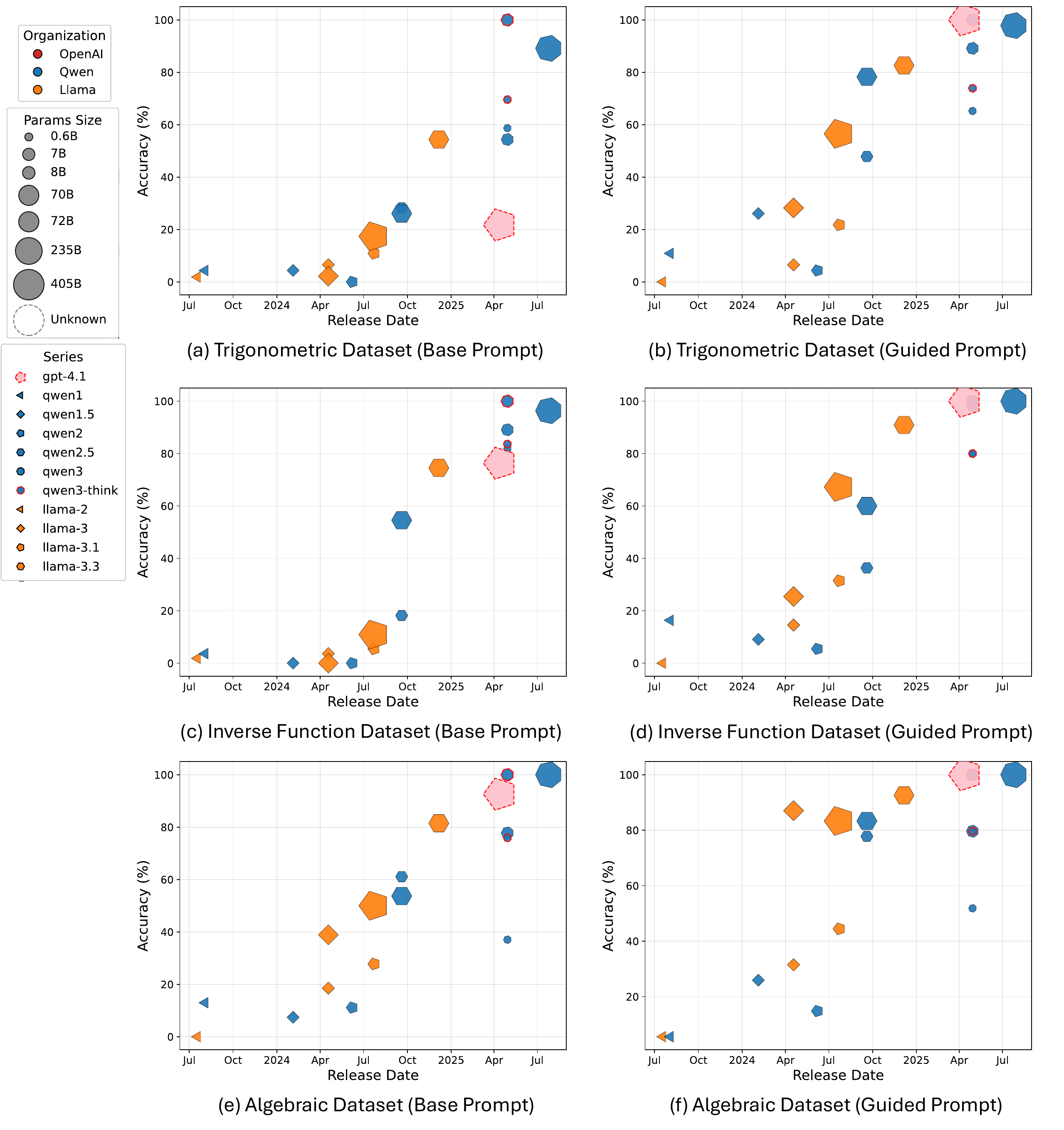}
      \caption{Evaluation of LLMs across varying release dates and model sizes.
(a), (c) and (e) present results on the trigonometric, inverse function and algebraic tasks using the base prompt, while (b), (d) and (f) present results on the same tasks using the guided prompt. }
  \label{fig:mislead_model_eval}
\end{figure}

\clearpage

\section{Empirical Study on Explicit/Implicit Solver}\lblapp{empirical_study_solver}

In this section, we present additional results on explicit versus implicit solvers.


\begin{figure}[h]
  \centering
  \begin{minipage}[b]{0.9\textwidth}
    \centering
    \includegraphics[width=\textwidth]{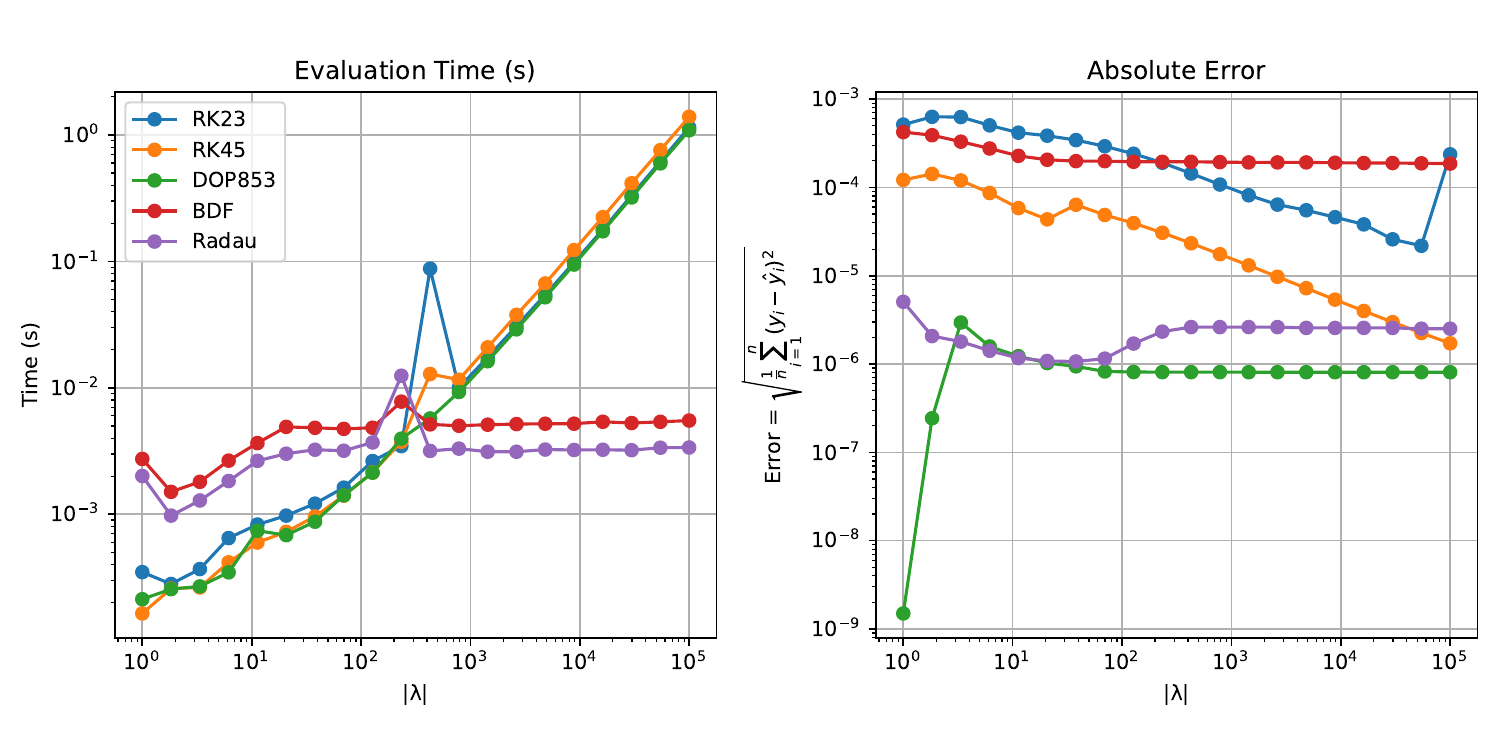}
    \caption*{(a) First-order ODE Experiemnt.}
  \end{minipage}
\vspace{-0.25em} 

  \begin{minipage}[b]{0.9\textwidth}
    \centering
    \includegraphics[width=\textwidth]{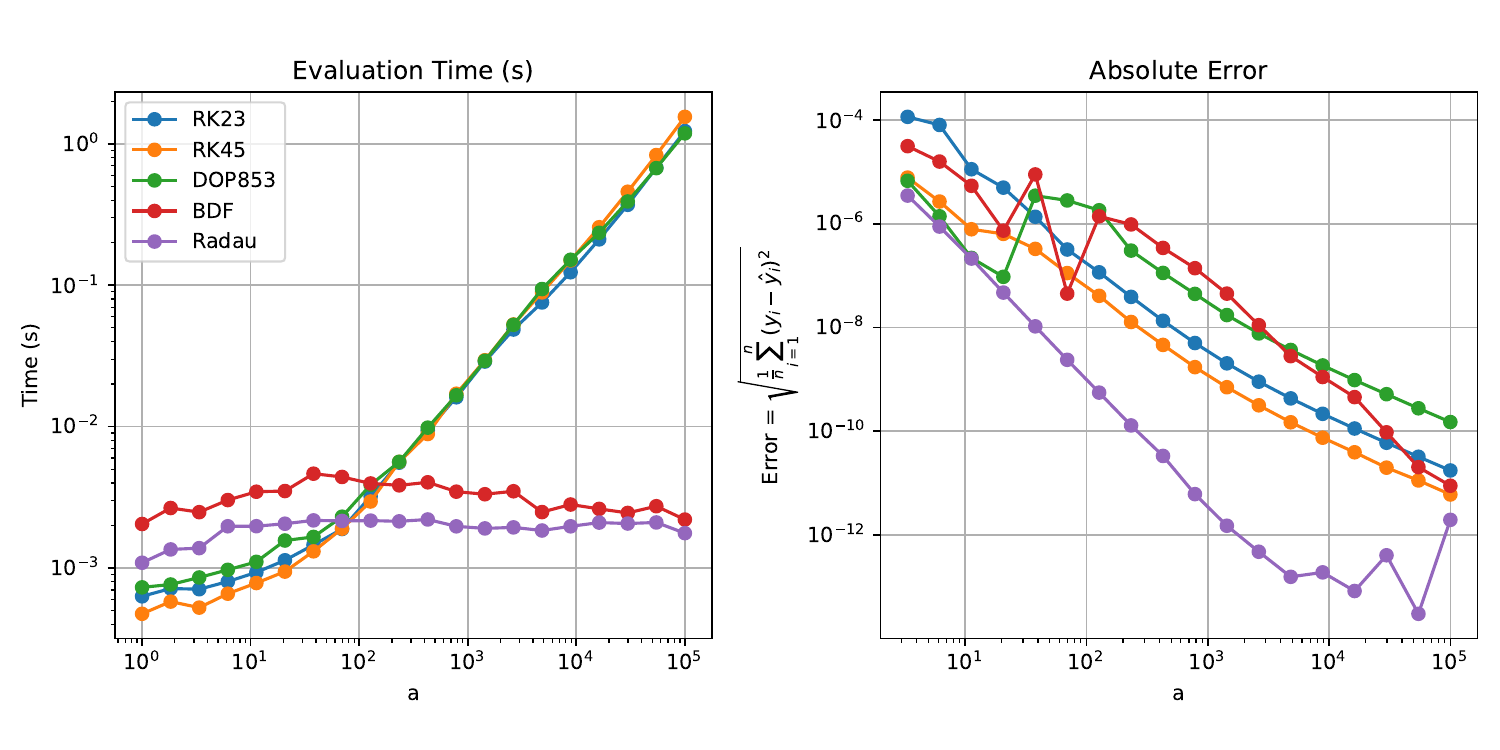}
    \caption*{(b) Second-order ODE Experiemnt.}
  \end{minipage}
  \caption{Empirical comparison of explicit and implicit solvers under varying stiffness. 
(a) Solver performance on the first-order ODE \( \frac{dy}{dt} = \lambda y \): left—runtime; right—RMSE. (b) Solver performance on the second-order ODE \( \frac{d^2y}{dt^2} + a \frac{dy}{dt} + y = 0 \): left—runtime; right—RMSE.
}
\label{fig:explicit_implicit_solver}
\end{figure}

\clearpage
\section{Failure Cases from Pattern Matching}\lblapp{mislead_failure_case}
In this section, we present reasoning traces showing how GPT-4.1 falls into pattern matching when determining the stiffness of ODEs across the trigonometric, inverse function, and algebraic subsets. We also include some successful cases in which large language models perform the correct mathematical reasoning and arrive at the correct result.

\subsection{Trigonometric Subset}
\noindent \textbf{1. Failed Case}
\begin{boxA}
\begin{itemize}
    \item \texttt{ODE example}: dy/dt = (50000*(cos(2*t) - 2*cos(t)**2 + 1))*y + sin(2*t),
    \item \texttt{Traces of Failed Reasoning}: This ODE has the derivative term dy/dt involving a large coefficient (50000) multiplied by y and trigonometric functions of t. Such a large coefficient creates rapid changes in y with respect to t and leads to the presence of both fast and slow dynamics. This multiscale behavior is characteristic of stiff ODEs, where explicit methods would require extremely small timesteps for stability, making them inefficient. Therefore, an implicit solver is more appropriate.
\end{itemize}
\end{boxA}

\noindent \textbf{2. Successful Case}
\begin{boxA}
\begin{itemize}
    \item \texttt{ODE example}: dy/dt = (20000*(sin(t)**2 + cos(t)**2) - 20000)*y + cos(t),
    \item \texttt{Traces of Successful Reasoning}: The ODE simplifies $(sin(t)^2 + cos(t)^2 = 1)$, so dy/dt = (20000*1 - 20000)*y + cos(t) = cos(t). This reduces to dy/dt = cos(t), a very simple, non-stiff ODE with no dependence on the large coefficient (since it cancels out). There are no rapidly decaying or strongly coupled components in the equation; the solution is smooth and varies on the scale of cos(t), not 20000. Therefore, stiffness is not an issue.
\end{itemize}
\end{boxA}

\subsection{Inverse Function Subset}
\noindent \textbf{1. Failed Case}
\begin{boxA}
\begin{itemize}
    \item \texttt{ODE example}: dy/dt = (4e7 * (sin(arcsin(t/2)) - t/2))*y + exp(t),
    \item \texttt{Domain}: t $\in$ [0, 1], 
    \item \texttt{Traces of Failed Reasoning}: The ODE's coefficient of y contains a large constant factor (4e7) that multiplies a function of t, which can lead to very rapid changes in the solution (potentially large negative or positive eigenvalues in the Jacobian). This is characteristic of stiff ODEs, as explicit methods may require prohibitively small step sizes for stability over even a short interval. Thus, an implicit solver is appropriate.
\end{itemize}
\end{boxA}

\noindent \textbf{2. Successful Case}
\begin{boxA}
\begin{itemize}
    \item \texttt{ODE example}: dy/dt = (1e7 * log(exp(t + 1)) - 1e7 * (t + 1))*y + cos(t),
    \item \texttt{Domain}: t $\in$ [0, 1], 
    \item \texttt{Traces of Successful Reasoning}: The log(exp(t + 1)) term equals t + 1, so 1e7 * log(exp(t + 1)) - 1e7 * (t + 1) cancels to 0. The Jacobian with respect to y is 0, so there are no stiff eigenvalues and the system is not stiff. Only the cos(t) term remains, thus an explicit solver is appropriate.
\end{itemize}
\end{boxA}

\subsection{Algebraic Subset}
\noindent \textbf{1. Failed Case}
\begin{boxA}
\begin{itemize}
    \item \texttt{ODE example}: dy/dt = ((t-5)**4 - t**4 + 20*t**3 - 150*t**2 + 500*t - 625)*y + sqrt(t+3),
    \item \texttt{Traces of Failed Reasoning}: To determine stiffness, let's analyze the ODE $dy/dt = ((t-5)^4 - t^4 + 20*t^3 - 150*t^2 + 500*t - 625)*y + sqrt(t+3)$. The right side shows linear dependence on y with a time-varying coefficient $P(t) = (t-5)^4 - t^4 + 20t^3 - 150t^2 + 500t - 625$. We should check for stiffness by looking for rapidly changing or large negative real parts in the eigenvalue (which, for linear scalar ODEs, is just P(t)). Evaluating P(t) at the endpoints: P(0) = 625, P(5) = (-625) + (2500) - (3750) + (2500) - (750) - (625) = large negative value. So, P(t) varies from positive to large negative within a short interval, potentially becoming very large in magnitude. Such large coefficients can make the ODE stiff, as the solution can rapidly decay or grow depending on sign and magnitude, and explicit solvers may need exceedingly small timesteps for stability.
\end{itemize}
\end{boxA}

\noindent \textbf{2. Successful Case}
\begin{boxA}
\begin{itemize}
    \item \texttt{ODE example}: dy/dt = ((t+5)**3 - t**3 - 15*t**2 - 75*t - 125)*y + sqrt(t+2),
    \item \texttt{Traces of Successful Reasoning}: To assess stiffness, examine the ODE's right-hand side: $dy/dt = ((t+5)^3 - t^3 - 15t^2 - 75t - 125)*y + sqrt(t+2)$. The coefficient of y simplifies as follows: $(t+5)^3 - t^3 - 15t^2 - 75t - 125 = (t^3 + 15t^2 + 75t + 125) - t^3 - 15t^2 - 75t - 125 = 0$. Therefore, dy/dt = sqrt(t+2). This is a simple function of t, and there is no y dependency beyond the initial value; thus, the ODE is not stiff. Its explicit integration is straightforward and does not require an implicit solver.
\end{itemize}
\end{boxA}

\clearpage


\end{document}